\documentclass{opt2020} 

\DeclareMathOperator*{\argmin}{\arg \min}

\newcommand{\dom}{\text{dom}}

\title[Generalised Perceptron Learning]{Generalised Perceptron Learning}

\optauthor{
 \Name{Xiaoyu Wang} \Email{xw343@cam.ac.uk}\\
 \addr Department of Applied Mathematics and  Theoretical Physics, University of Cambridge
 \AND
 \Name{Martin Benning} \Email{m.benning@qmul.ac.uk}\\
 \addr School of Mathematical Sciences, Queen Mary University of London
}

\begin{document}

\maketitle
\begin{abstract}%
We present a generalisation of Rosenblatt's traditional perceptron learning algorithm to the class of proximal activation functions and demonstrate how this generalisation can be interpreted as an incremental gradient method applied to a novel energy function. This novel energy function is based on a generalised Bregman distance, for which the gradient with respect to the weights and biases does not require the differentiation of the activation function. The interpretation as an energy minimisation algorithm paves the way for many new algorithms, of which we explore a novel variant of the iterative soft-thresholding algorithm for the learning of sparse perceptrons.
\end{abstract}


\section{Introduction}


In this work, we consider the problem of training perceptrons with (potentially) non-smooth activation functions. Standard training procedures such as subgradient-based methods often have undesired properties such as potentially slow convergence \cite{polyak-book,bertsekas-book, chambolle2016introduction}. 
We revisit the training of an artificial neuron \cite{mcculloch1943logical} and further show that Rosenblatt's perceptron learning algorithm can be viewed as a special case of incremental gradient descent method (cf. \cite{bertsekas2011incremental}) with respect to a novel choice of energy. This new interpretation allows us to devise approaches that avoid the computation of sub-differentials with respect to non-smooth activation functions and thus provides better handling for the non-smoothness in the overall minimisation objective during training.




The paper is organised as follows. We begin with a recap of the perceptron model and perceptron learning algorithms in Section \ref{sec:perceptron}. Next, we introduce an energy function based on the generalised Bregman distance and demonstrate how Rosenblatt's learning algorithm can be interpreted as an incremental gradient method applied to this energy in Section \ref{sec:bregman-loss}. In Section \ref{sec:application-numerical} we present numerical results for learning sparse perceptrons and compare the results to those obtained from subgradient-based methods. We conclude and give an outlook of future research directions in Section \ref{sec:conclusion}.

\section{Perceptrons revisited}\label{sec:perceptron}

A (generalised) perceptron can be considered as an artificial neuron \cite{mcculloch1943logical} or one-layer feed forward neural network of the form
\begin{align}
    y = \sigma\left( W^\top x + b \right) \, .\label{eq:perceptron}
\end{align}
Here $\sigma$ denotes the (point-wise) activation function, $W \in \mathbb{R}^{m \times n}$ is the weight-matrix and $b \in \mathbb{R}^n$ is the bias-vector. The vector $x \in \mathbb{R}^m$ and the vector $y \in \mathbb{R}^n$ denote the input, respectively the output, of the perceptron. The first perceptron learning algorithm was proposed by Frank Rosenblatt in 1957 \cite{rosenblatt1957perceptron} and is summarised in Algorithm \ref{alg:perceptron-learning}, where $s$ denotes the number of training samples. This early work studies perceptrons in the context of binary supervised classification problems and uses the Heaviside step function as the activation function $\sigma$ to generate binary output.

\begin{algorithm2e}[!t]\label{alg:perceptron-learning}
Initialize $W^0$, $b^0$\;
\For{$k = 1, 2, \ldots$}{
    \For{$i = 1, \ldots, s$}{
        $e_i = y_{i} - \sigma\left( (W^k)^\top x_i + b^k\right)$\\
        $W^{k+1} = W^k +  e_i \, x_i^\top$ \\
        $b^{k + 1} = b^k + e_i$
  }
  }
\label{rosenblatt}
\caption{Rosenblatt's Perceptron Learning Algorithm}
\end{algorithm2e}

Alternatively, the problem of training the weights and biases of a perceptron of the form of Equation \eqref{eq:perceptron} can be formulated as a minimisation problem of the form
\begin{align}
    \min_{W, b}F(W,b) :=  \frac{1}{s} \sum_{i}^{s} L\left(y_i \, , \, \sigma(W^\top x_i + b) \right) + \alpha R(W, b) \, ,\label{eq:minimisation-problem}
\end{align}
where $F$ is the overall cost- or loss-function and $L$ is the data function that is chosen a-priori (usually based on prior assumptions of the statistical distribution of the data). The function $R$ is a regularisation function that allows to encode prior information on weight and bias which can be useful to combat ill-conditioning of the data matrix \cite{benning2018} and help to control the validation error \cite{goodfellow2016deep,shalev2014understanding}. Both terms are balanced with a positive regularisation parameter $\alpha$. 

The overall objective function in Equation \eqref{eq:minimisation-problem} is usually minimised by a gradient- or subgradient-based algorithm such as (sub-)gradient descent. If we for instance choose $R \equiv 0$ and $L(y, \sigma(W^\top x + b)) = \frac{1}{2} \| y - \sigma(W^\top x + b) \|^2$ and minimise \eqref{eq:minimisation-problem} via mini-batch subgradient descent, we obtain the algorithm described in Algorithm \ref{alg:mse-incremental-subgradient-descent}, where $B_k \subset \{1,2,\dots s\}$ is a (mini-)batch of indices chosen at iteration $k$. When $|B_k| = 1$, this corresponds to the incremental or stochastic subgradient descent method. Here, $\sigma^\prime$ denotes the derivative of $\sigma$, respectively a subderivative if $\sigma$ is not differentiable. Hyper-parameters $\tau_w^k > 0$ and $\tau^k_b > 0$ denote the
learning rates at iteration $k$. 





We will reveal in the following section that the original Rosenblatt's learning algorithm is in fact a special case of incremental gradient descent method with respect to a novel choice of energy.


\section{Perceptron training: minimising a Bregman loss function}\label{sec:bregman-loss}

We replace the data term in Equation \eqref{eq:minimisation-problem} with a loss function that enables us to interpret Algorithm \ref{alg:perceptron-learning} as an incremental gradient descent method for a special class of activation functions: so-called proximal maps \cite{moreau1962fonctions}.  

\begin{definition}[Proximal map]
The proximal map $\sigma:\mathbb{R}^n \rightarrow \text{dom}(\Psi) \subset \mathbb{R}^n$ of a proper, lower semi-continuous and convex function $\Psi:\mathbb{R}^n \rightarrow \mathbb{R} \cup \{ \infty \}$ is defined as
\begin{align*}
    \sigma(z) := \argmin_{u \in \mathbb{R}^n} \left\{ \frac12 \| u - z \|^2 + \Psi(u) \right\} \, .
\end{align*}
\end{definition}
\begin{example}[Rectifier]\label{ex:prox-maps}
There are numerous examples of proximal maps, see e.g. \cite{combettes2020deep}. In particular, the rectifier or ramp function can be interpreted as the proximal map of the characteristic function over the non-negative orthant:
\begin{align*}
    \Psi(u) := \begin{cases} 0 & u \in [0, \infty)^n \\ \infty & \text{otherwise} \end{cases} \qquad &\implies \qquad \sigma(z)_j = \max( 0, z_j ) \, , \; \forall j \in \{1, \ldots, n \} \, ,
\end{align*}
\end{example}

\begin{algorithm2e}[!t]\label{alg:mse-incremental-subgradient-descent}
Initialize $W^0$ and $b^0$ \;
\For{$k = 1, 2, \ldots$}{
    Choose $B_k \subset \{1,2,\dots s\}$ either at random or deterministically.\\
    $g_w^{k} = \frac{1}{|B_k|} \sum_{i \in B_k} [\sigma((W^k)^\top x_i + b^k) - y_i] \sigma^\prime((W^k)^\top x_i + b^k) \, x_i^\top $ \\
    $W^{k+1} = W^k -  \tau_w^k \, g_w^k  $ \\
    $g_b^k = \frac{1}{|B_k|} \sum_{i \in B_k} [\sigma((W^k)^\top x_i + b^k) - y_i] \sigma^\prime((W^k)^\top x_i + b^k)$\\
    $b^{k+1} = b^k - \tau_b^k \; g_b^k $ \\
    
  }
\label{subggd}
\caption{Mini-batch Subgradient Descent}
\end{algorithm2e}

The loss function that we propose in this work is based on a concept known as the (generalised) Bregman distance \cite{bregman1967relaxation,censor1981iterative,eckstein1993nonlinear,kiwiel1997free}, which is defined as follows.
\begin{definition}[Generalised Bregman distance]
The generalised Bregman distance of a proper, lower semi-continuous and convex function $\Phi$ is defined as
\begin{align*}
    D_\Phi^{q(v)}(u, v) := \Phi(u) - \Phi(v) - \langle q(v), u - v \rangle \, .
\end{align*}
Here, $q(v) \in \partial \Phi(v)$ is a subgradient of $\Phi$ at argument $v \in \mathbb{R}^n$ and $\partial \Phi$ denotes the subdifferential of $\Phi$.
\end{definition}

Inspired from \cite{geiping2019parametric} and based on the definition of the generalised Bregman distance and the assumption $y \in \dom(\Psi)$, we propose the data term
\begin{align}
    L(y, \sigma(z)) := \frac12 \| y - \sigma(z) \|^2 + D_\Psi^{z - \sigma(z)}(y, \sigma(z)) \, ,\label{eq:bregman-loss}
\end{align}
for the (valid) subgradient $z - \sigma(z) \in \partial \Psi(\sigma(z))$. The motivation in using Equation \eqref{eq:bregman-loss} as a loss function (instead of only using the squared two norm) lies in the simplicity of its gradient.
\begin{theorem}
For fixed $y \in \dom(\Psi)$, the gradient of $L$ as defined in Equation \eqref{eq:bregman-loss} with respect to the argument $z$ reads
\begin{align*}
    \nabla_z L(y, \sigma(z)) = \sigma(z) - y \, .
\end{align*}
\end{theorem}
The proof for this theorem is given in the appendix.

\begin{algorithm2e}[!t]\label{alg:rosenblatt-ista}
Initialize $W^0$ and $b^0$ \;
\For{$k = 1, 2, \ldots$}{
    Choose $B_k \subset \{1,2,\dots s\}$ either at random or deterministically.\\
    $g_w^k = \frac{1}{|B_k|} \sum_{i \in B_k} [\sigma((W^k)^\top x_i + b^k) - y_i ] \, x_i^\top $ \\
    $W^{+}  = W^k -  \tau_w^k \,g_w^k $ \\
    $W^{k+1} = \argmin_{W \in \mathbb{R}^{m \times n}} \{ \frac12 \| W^{+} - W\|^2 + \alpha \|W\|_1\}$ \\
    $g_b^k = \frac{1}{|B_k|} \sum_{i \in B_k} [\sigma((W^k)^\top x_i + b^k) - y_i ]$\\
    $b^{k+1} = b^k - \tau_b^k \, g_b^k$ \\
    
  }
\label{subggd}
\caption{Rosenblatt-ISTA Algorithm}
\end{algorithm2e}

If we use the data term defined in Equation \eqref{eq:bregman-loss} in the Minimisation Problem \eqref{eq:minimisation-problem} and apply an incremental gradient descent strategy with constant step-size or learning rate one, we immediately obtain Rosenblatt's learning algorithm as defined in Algorithm \ref{alg:perceptron-learning} as a consequence of the chain rule. In other words, \textbf{an incremental or stochastic gradient descent method applied to the Bregman loss \eqref{eq:bregman-loss} generalises Rosenblatt's original learning algorithm to a broader class of (proximal) activation functions, for which Algorithm \ref{alg:perceptron-learning} can be interpreted as an energy minimisation method}, just with the different Energy \eqref{eq:bregman-loss}. We also note that as a special case when the function $\Psi$ is simply zero, the activation function $\sigma$ is the identity function and \eqref{eq:bregman-loss} is equivalent to the squared loss $\frac{1}{2}\|y-W^\top x\|_2^2$. Hence, incremental gradient descent applied to this energy is equivalent to the Adaline delta rule, first proposed by Bernard Widrow in 1960 \cite{widrow1960adaptive}.

\textbf{The beauty of interpreting Rosenblatt's learning algorithm as an energy minimisation problem is that we can apply many other algorithmic strategies to the same energy}. In the following section, we demonstrate how we can easily set up an iterative soft-thresholding generalisation of the Rosenblatt algorithm to train a perceptron with sparse weights without having to differentiate the activation function.

\section{Application and numerical experiment} \label{sec:application-numerical}



In this section, \textbf{we use the energy minimisation interpretation to develop a new way of learning sparse perceptrons} (cf. \cite{jackson1996learning}), \textbf{without having to differentiate the activation function}. We consider a LASSO-type cost function \cite{tibshirani1996regression}, i.e. we choose the regularisation term $R$ in Equation \eqref{eq:minimisation-problem} to be $R(W, b) = \|W\|_1$ in order to promote sparsity of the weight matrix. Here, $\| \cdot \|_1$ denotes the $\ell^1$-norm. We consider using a rectifier activation function as in Example \ref{ex:prox-maps}. 
To avoid having to differentiate the non-smooth activation function when solving Minimisation Problem \eqref{eq:minimisation-problem}, we propose to use the Bregman Data Term \eqref{eq:bregman-loss} as choice for $L$ in combination with the Iterative Soft-Thresholding Algorithm (ISTA) \cite{daubechies2004iterative} to handle the non-smooth $\ell^1$ norm. We name the resulting algorithm \textbf{Rosenblatt-ISTA}, which is summarised in Algorithm \ref{alg:rosenblatt-ista}. In numerical experiments, we choose $|B_k| = s$, which corresponds to using the full (sub-)gradient. We compare our approach with three baseline approaches, which aim to solve \eqref{eq:minimisation-problem} with a squared Euclidean loss instead. The first approach uses a standard subgradient method with constant step size, where we define the subderivative $\sigma^\prime$ of the rectifier activation function at zero to take on value one. The second approach uses a subgradient method with diminishing step size $\tau_w^k = \frac{\tau_w^0}{\sqrt{k}}$, which guarantees convergence but gives slower convergence. The third method also follows ISTA but instead performs a subgradient descent step in the direction of the (now) non-smooth data term $L$ during each iteration.

\begin{figure}[!t]
    \centering
    \includegraphics[scale= 0.25]{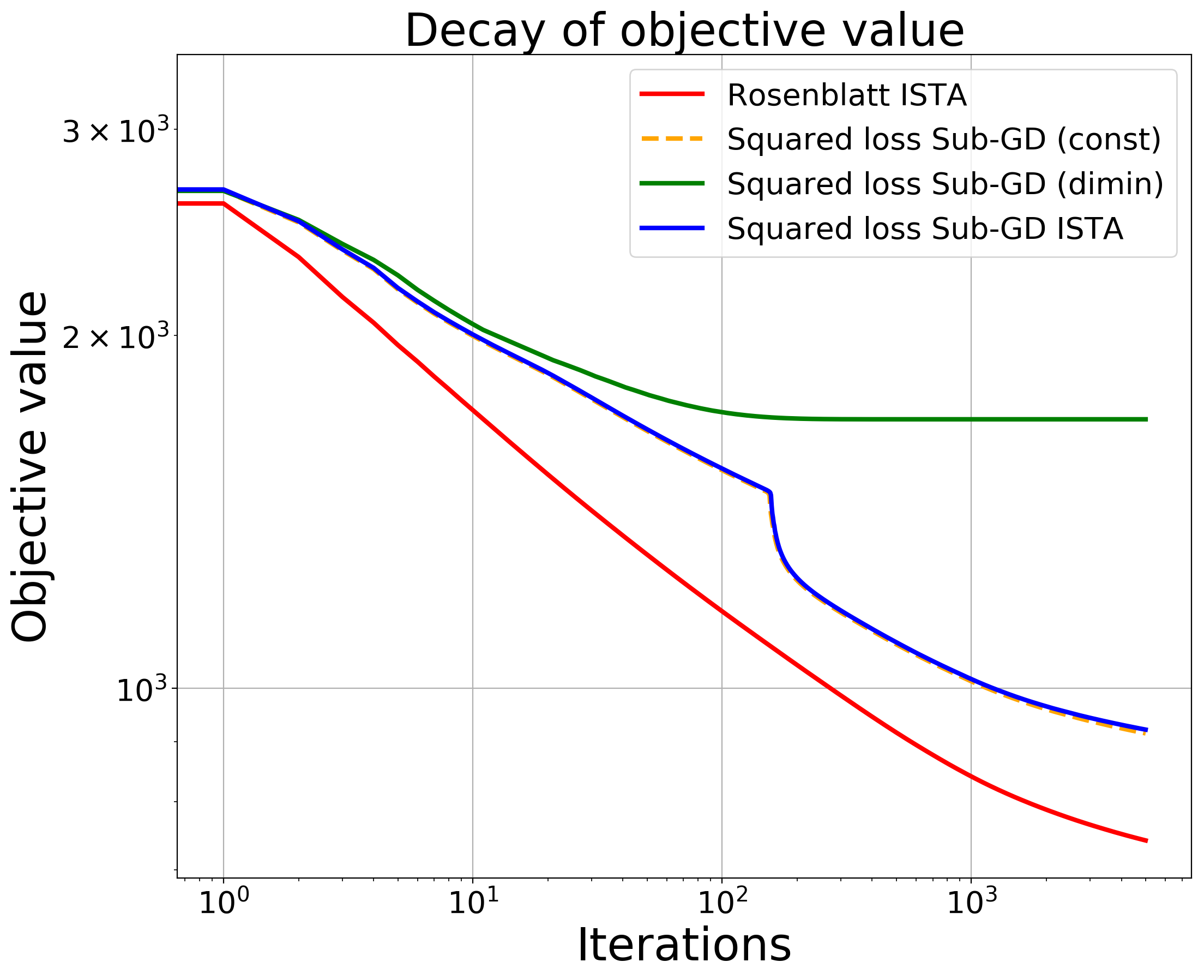}
    \includegraphics[scale= 0.25]{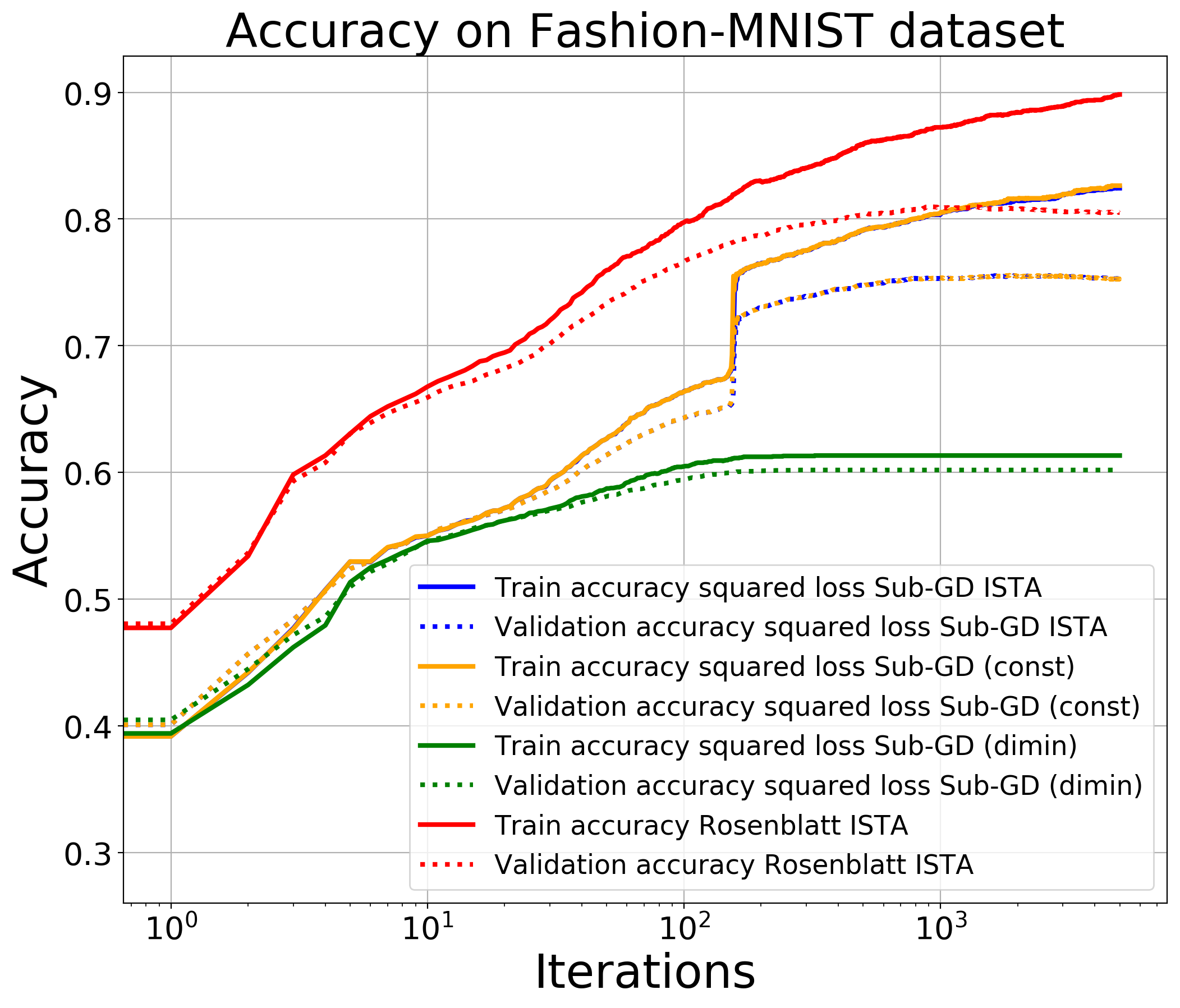}
    \caption{We compare our method of training sparse perceptrons with three baseline approaches as described in Section \ref{sec:application-numerical}. The plot on the left shows the decay of objective value, i.e. Bregman loss plus $\ell^1$ regularisation for our proposed scheme and the squared Euclidean loss plus $\ell^1$ regularisation for the baseline approaches. The plot to the right shows the change of training and validation accuracy over the number of iterations. All experiments are performed on the Fashion-MNIST dataset \cite{xiao2017fashion}. 
    }
    \label{fmnist}
\end{figure}

We test the algorithm performances with a toy classification example based on images from the Fashion-MNIST dataset \cite{xiao2017fashion} with the four schemes described earlier. All methods and results are implemented in Python with the Numpy library. We train the perceptrons on 3,000 images selected from the training dataset and validate them on 10,000 images from the test dataset. The regularisation parameters $\alpha$ are set to 0.9 for our proposed objective and 0.81, 0.81, 0.85 for the three baseline methods respectively. The choices of $\alpha$ are determined via cross validation. All four approaches are initialised with the same weight, bias and step size. The overall objective value and accuracy results are visualised in Figure \ref{fmnist}. We can see that our proposed training scheme is able to converge faster than the baseline approaches, and also achieves higher training and validation accuracy. It is important that these results are just a proof-of-concept and do not aim at outperforming more complicated neural network architectures.

\section{Conclusion and future work}\label{sec:conclusion}
In this work, we discussed the learning of a generalised perceptron model with proximal activation functions. We demonstrated that, from an energy minimisation point of view, the original Rosenblatt's perceptron learning algorithm can be interpreted as an incremental gradient method with respect to a novel choice of data term, based on a generalised Bregman distance. In addition, this interpretation also generalises the classical Adaline delta rule. 


We have modified this interpretation to learn a sparse perceptron with Rosenblatt-ISTA. We showed numerical results for which our approach outperforms three other subgradient-based baseline approaches in terms of achieving both faster convergence and higher accuracy results. A future direction of this work is to extend the proposed generalised perceptron learning scheme to more complex architectures such as multi-layer perceptrons or convolutional neural networks. 




\bibliography{sample}

\newpage
\clearpage
\appendix

\section{Proof of Theorem 3 in Section \ref{sec:bregman-loss}}

Let $E_z(\sigma(z))$ denote the Moreau-Yosida Regularisation \cite{moreau1965proximite,yosida1964functional} of $\Psi$, where $E_z$ is defined as
\begin{equation*}
    E_z(x) := \frac{1}{2}\|x-z\|^2 + \Psi(x) \,.
\end{equation*}
We then observe
\begin{align*}
    L(y,\sigma(z)) &= \frac12 \| y - \sigma(z) \|^2 + D_\Psi^{z - \sigma(z)}(y, \sigma(z))  \\
    &= \frac12 \| y - \sigma(z) \|^2 + \Psi(y) - \Psi(\sigma(z)) - \langle z-\sigma(z), y - \sigma(z) \rangle \\
    &= \frac12 \langle y-\sigma(z), y-\sigma(z) \rangle - \langle z-\sigma(z), y - \sigma(z) \rangle + \Psi(y) - \Psi(\sigma(z)) \\
    &= \frac12 \langle y-z, y-\sigma(z) \rangle + \frac12 \langle \sigma(z)-z , y - \sigma(z) \rangle + \Psi(y) - \Psi(\sigma(z)) \\
    &= \frac12 \langle y-z, y-z - \sigma(z)+z \rangle - \frac12 \langle \sigma(z)-z , \sigma(z) - y \rangle + \Psi(y) - \Psi(\sigma(z)) \\
    &= \frac12 \langle y-z, y-z \rangle - \frac12 \langle \sigma(z)-y, \sigma(z) - z\rangle - \frac12 \langle y-z , \sigma(z) - z \rangle + \Psi(y) - \Psi(\sigma(z)) \\
    &= \frac12 \|y-z\|^2 - \frac12 \langle \sigma(z)-y + y - z, \sigma(z) - z \rangle + \Psi(y) - \Psi(\sigma(z)) \\
    &= \frac12 \|y-z\|^2 + \Psi(y) - \frac12 \|\sigma(z)- z\|^2 - \Psi(\sigma(z)) \\
    &= E_z(y) - E_z(\sigma(z)) \,.
\end{align*}
It is well known (cf. \cite{parikh2014proximal}) that the gradient of the Moreau-Yosida regularisation satisfies
\begin{align*}
     \nabla E_z(\sigma(z)) = z - \sigma(z) \, .
\end{align*}
Therefore, we derive the gradient of $L$ with respect to the argument $z$ as
\begin{align*}
    \nabla_z L(y,\sigma(z)) &= \nabla_z E_z(y) - \nabla_z E_z(\sigma(z)) \\
    &= z - y - z + \sigma(z) \\
    &= \sigma(z) - y \,.
\end{align*}
This concludes the proof.\hfill\rlap{$\square$}

%


\end{document}